\documentclass[sigconf,screen]{acmart}

\usepackage{soul}
\usepackage{url}
\usepackage{hyperref}
\usepackage{graphicx}
\usepackage{amsmath}
\usepackage{algorithm}
\usepackage[noend]{algorithmic}
\usepackage{wrapfig}
\usepackage{xspace,color}
\usepackage{enumitem}
\usepackage{subcaption}
\usepackage{graphics}
\usepackage{amsfonts}       
\usepackage{xcolor}
\usepackage{multirow}
\usepackage{nicefrac}       
\usepackage{microtype}      
\usepackage{makecell}
\theoremstyle{definition}

\newtheorem{definition}{Definition}
\newtheorem{task}{Task}

\setlist[itemize]{leftmargin=*}
\setlist[enumerate]{leftmargin=*}
\setlist{nosep}

\newcommand{\method}{\texttt{TREEMENT}\xspace}

\acmConference[BCB '23]{Conference on Bioinformatics, Computational Biology, and Health Informatics}{September 03--06,
  2023}{Houston, TX}

\begin{document}

\title[TREEMENT: Interpretable Patient-Trial Matching via Tree-Based Memory Network]{TREEMENT: Interpretable Patient-Trial Matching via Personalized Dynamic \underline{Tree}-Based \underline{Me}mory \underline{N}e\underline{t}work}

\author{Brandon Theodorou$^{1}$, Cao Xiao$^{2}$, and Jimeng Sun$^{1}$}
\affiliation{
\institution{University of Illinois at Urbana-Champaign$^{1}$}
\country{}}
\affiliation{
\institution{Relativity Inc.$^{2}$}
\country{}}

\begin{abstract}
Clinical trials are critical for drug development but often suffer from expensive and inefficient patient recruitment. In recent years, machine learning models have been proposed for speeding up patient recruitment via automatically matching patients with clinical trials based on longitudinal patient electronic health records (EHR) data and eligibility criteria of clinical trials. However, they either depend on trial-specific expert rules that cannot expand to other trials or perform matching at a very general level with a black-box model where the lack of interpretability makes the model results difficult to be adopted.

To provide accurate and interpretable patient trial matching, we introduce a personalized dynamic tree-based memory network model named \method. It utilizes hierarchical clinical ontologies to expand the personalized patient representation learned from sequential EHR data, and then uses an attentional beam-search query learned from eligibility criteria embedding to offer a granular level of alignment for improved performance and interpretability. We evaluated \method against existing models on real-world datasets and demonstrated that \method outperforms the best baseline by 7\% in terms of error reduction in criteria-level matching and achieves state-of-the-art results in its trial-level matching ability. Furthermore, we also show \method can offer good interpretability to make the model results easier for adoption.
\end{abstract}

\maketitle

\section{Introduction}
Clinical trials, where patients are enrolled to evaluate the safety and efficacy of new medical or surgical interventions, are essential for the development of new treatments. However, they often struggle with expensive and inefficient patient recruitment as finding enough patients who meet all eligibility criteria is nontrivial~\cite{feller2015one}. For example, more than 50\% of trials face delays due to insufficient recruitment~\cite{hargreaves2016clinical}. In addition, inaccurate recruitment will also cause the recruitment cost to be tripled, consequently affecting the treatment cost and adding more burden to patients~\cite{mdgroup_2020}.

Automating patient-trial matching using longitudinal patient electronic health records (EHR) data and eligibility criteria (EC) of clinical trials brings the promise to alleviate the pain of clinical trial recruitment. Over the past years, machine learning algorithms were proposed to address the clinical trial recruitment challenges, including both rule-based and deep learning based approaches.  \underline{Rule-based approaches} perform patient retrieval either based on expert annotations~\cite{weng2011elixr,kang2017eliie} or by training supervised learning classifiers to extract rules~\cite{Criteria2Query}. 
While they were highly interpretable, they typically failed to perform well due to  the significant variance and overlap in medical terminology and the inflexible nature of rule-based systems.  \underline{Deep learning approaches} such as~\cite{DeepEnroll,COMPOSE} instead embed EC sentences and patient EHRs into a shared latent space and match them based on semantic similarity. 
They generally have improved accuracy in patient-trial matching but could not offer the same sufficient level of interpretability. 
For example, while some explanation can be obtained via  model weights or attention mechanisms, the attentions are designed to focus on abstract disease concept levels (e.g., level 1 diagnosis) rather than granular concepts (e.g., Type 2 diabetes without complication), which makes the results overly general.

To summarize, we identify two open challenges.
\begin{enumerate}
    \item \textbf{Challenge in identifying patients with more granular medical concepts}. The eligibility criteria for clinical trials need to be accurately mapped to granular medical codes. While patients can have an incredibly diverse set of medical codes in their history. Given the large global vocabulary of medical codes, it is infeasible to have static memory slots for each individual concept. To bypass this, existing works such as ~\cite{COMPOSE} proposed to reduce the memory dimension by aggregating these codes into higher-level concepts. This strategy is ineffective as diverse medical codes will share the same memory slots, and the resulting aggregated code embeddings are insufficient for accurate patient identification. 
    
    \item \textbf{Challenge in learning interpretable patient trial matching}. Oftentimes, the matching is determined not just by a specific medical code but instead by what a code signifies through accompanying higher or lower-level information. It is essential to have a granular understanding of the matching. For example, a patient may have a basal cell carcinoma code in their record, but the criteria may refer to a more general skin cancer diagnosis that they also possess. Thus, this problem brings an additional challenge of expanding and infusing representations with ontological information and the need to handle codes whose higher-level categorization may overlap.\\
\end{enumerate}

This paper presents \method to address these challenges. \method takes patient EHR and trial EC sentences as input data and outputs a personalized and interpretable prediction of whether patients are fit for a particular clinical trial, enabled by the following technical contributions:

\begin{enumerate}
    \item \textbf{A dynamic tree-based memory network for efficient personalized patient data embedding}. To capture multiple concept granularity while mitigating the challenge caused by the large size of medical codes, we design a dynamic tree-based memory network to represent a patient based on their personal longitudinal and hierarchical medical history. For each patient, the memory tree  only consists of memory slots for the  medical concepts in the patient's record. It is also granular, with an individual slot for each concept, joined and made into a tree through shared higher-level hierarchical concepts. Such a design allows for efficient embedding  and provides the proper concept granularity for model interpretation.
    \item \textbf{Attentional beam search over knowledge-enhanced embedding for interpretable matching}. We design an attentional beam search method for efficient top-down exploration of patients' knowledge-enhanced memory trees based on the criteria embedding query. The search outputs the prediction and an interpretable tree structure to explain and verify its predictions.\\

\end{enumerate}
We evaluated \method against state-of-the-art models  and demonstrated that \method outperforms the best baseline in criteria and trial level matching by up to 7\% in terms of error reduction in accuracy and F1 scores. Furthermore, we  present the strong interpretability improvement of \method via case studies in which we show that \method correctly identifies patient's most relevant medical codes for  easy verification of its prediction.

\section{Related work}

\noindent\textbf{Patient-trial matching} refers to the task of matching patients with clinical trials based on the concept similarity encoded by the patient and trial data. Both rule-based approaches and deep learning based approaches were proposed in the past. Rule-based approaches rely on expert annotation or rules extracted using supervised learning. For example, \cite{weng2011elixr} proposed a syntactic mining and pattern matching framework to extract information from and label eligibility statements. \cite{bustos2018learning} performed similar text mining on criteria statements before undergoing representation learning on that mined information. \cite{alicante2016unsupervised} offered a similar entity and relation extraction setup but on patient records.  \cite{Criteria2Query} combined machine learning and rule-based methods to generate SQL queries for ECs. The rule based approaches often yield poor recall due to morphological variants and inadequate rule coverage~\cite{COMPOSE}. More recently, deep learning based methods such as \cite{DeepEnroll} and \cite{COMPOSE} jointly embed patient records and trial ECs in the same latent space, and then aligns them using attentive inference. They offered large improvements in performance, but the interpretability through attention to the broad categories they aligned to offered inferior interpretability than was necessary for human verification in a production system.\\

\noindent\textbf{Memory Networks} are a form of attention mechanism that supports knowledge extraction by building a structured set of memory cells that are then ``queried" through an input vector to attend only to the values in the relevant cells. 
Memory networks have been used in diverse cross-modality application areas such as text-based queries for image retrieval~\cite{huang2019acmm,chen2020imram}. In patient-trial matching task, ~\cite{COMPOSE} utilizes embedding criteria queries into a patient representation memory network on the task we currently consider. However, while memory networks offer the ability to align multiple modalities through the query and response, they can only align the query with whatever feature individual memory cells represent. Thus, if the cells are too broad, as was the case with ~\cite{COMPOSE} the alignment can be wasteful and the interpretability offered by the attention can be less granular than desired. 
To overcome those limitations, \method makes memory networks dynamic, granular, and hierarchical with a new tree-based memory to increase both granularity and interpretability. \footnote{\textbf{Note} There are several models named \textbf{tree memory networks}~ \cite{diehl2019tree,fernando2018tree}, however they are irrelevant to our model setting. To be specific, their goals were to improve long short-term memory networks for better modeling long-term dependencies via a fixed tree-based memory while we design tree-based memory cells organized by hierarchical medical ontologies to better capture the parent-child relationships of medical codes.}

\section{Problem Formulation}

\begin{definition}[\textbf{Patient EHR Data}] In this paper, we represent patient EHR data as $P = (\mathbf{E}, \mathbf{M}_p, \mathbf{d})$, where $\mathbf{E} \in \mathbb{R}^{n_v \times 3 \times n_c}$ is a sequence of $n_v$ visits with one-hot $n_c$-dimensional vectors for possible diagnosis, procedure, and medication codes at each visit, $\mathbf{M}_P \in \mathbb{R}^{n_v \times 3}$ represents the presence of each code at each visit, and $\mathbf{d} \in \mathbb{R}^{3}$ represents patients' demographic information such as gender and age.
\end{definition}

 \begin{definition}[\textbf{Clinical Trial EC Statement}] For clinical trial EC statement, the sentences and corresponding length through its mask are denoted as 
 $\mathbf{S} \in \mathbb{R}^{n_s \times n_w}$ and  $\mathbf{m}_C \in \mathbb{R}^{n_s}$, respectively, where $n_s$ is the max sentence length and $n_w$ is the number of tokens in the vocabulary such that a sentence is represented as a series of one-hot vectors for each token.
\end{definition}
 
 \begin{task}[\textbf{Patient Trial Matching}] The patient trial matching task considers patient data $P = (\mathbf{E}, \mathbf{M}_p, \mathbf{d})$ and EC sentence $C = (\mathbf{S}, \mathbf{m}_c)$ as input and classifies patient-criteria pairs into one of three categories of ``match," ``mismatch," and ``unknown" via a predicted probability distribution $\mathbf{\hat{y}} \in \mathbb{R}^3$ over the three classes against the true one-hot label vector $\mathbf{y} \in \mathbb{R}^3$.
 \end{task}
 \noindent We list relevant notations in Table \ref{tab:Notation} for reference.

\begin{table}
\centering
\caption{Table of Notations}
\begin{tabular}{c|l} \toprule
Notation   & Description \\ \midrule
$P$          & The patient \\
$C$          & The eligibility criteria \\
$\mathbf{y} \in \mathbb{R}^3$  & The patient-criteria label \\ \hline
$\mathbf{E} \in \mathbb{R}^{n_v \times 3 \times n_c}$ & The patient's electronic health record \\
$\mathbf{M}_P \in \mathbb{R}^{n_v \times 3}$ & The patient's EHR mask \\ 
$\mathbf{d} \in \mathbb{R}^{3}$ & The patient's demographics \\ 
$\mathbf{S} \in \mathbb{R}^{n_s \times n_w}$ & The criteria sentence \\ 
$\mathbf{m}_C \in \mathbb{R}^{n_s}$ & The criteria sentence mask \\ \hline
$n_v \in \mathbb{N}$  & The maximum number of visits \\
$n_c \in \mathbb{N}$  & The size of the code vocabulary \\
$n_s \in \mathbb{N}$  & The maximum criteria sentence length \\
$n_w \in \mathbb{N}$  & The size of the word vocabulary \\\bottomrule
\end{tabular}
\label{tab:Notation}
\end{table}

\section{The \method Model}

\begin{figure*}
    \centering
    \includegraphics[scale=0.41]{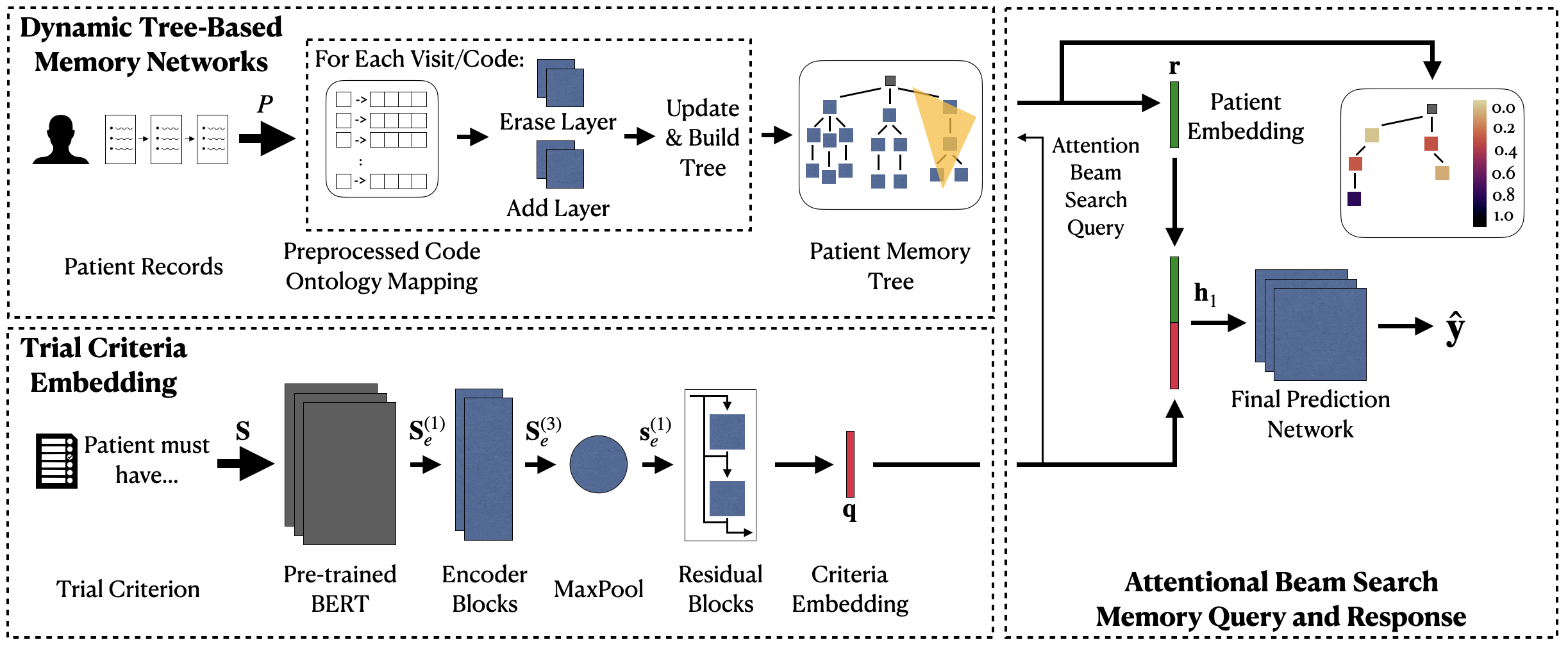}
    \caption{The \method model. We first utilize a patient's longitudinal health record and the underlying medical code ontology to build our personalized and dynamic memory tree containing specific medical codes in a patient's history enriched by their temporal and hierarchical context. Next, we embed a trial eligibility criteria sentence with a mix of pre-trained and trainable Transformer encoder layers, a MaxPool compression, and residual blocks to produce $\mathbf{q}$. We then utilize an attention beam search query on our memory tree using $\mathbf{q}$ to perform a targeted search of the patient's medical history to extract the relevant medical codes for the specific criterion without looking at any extraneous or irrelevant data. These relevant codes are combined in the query response vector $\mathbf{r}$. They also compose the interpretability output. Finally, we concatenate $\mathbf{q}$ and $\mathbf{r}$ to form $\mathbf{h}_1$ and use our final prediction network to generate \method's classification prediction $\hat{\mathbf{y}}$.}
    \label{fig:TREEMENTArchitecture}
\end{figure*}

In this section, we introduce  \method (illustrated in Figure ~\ref{fig:TREEMENTArchitecture}), which is a dynamic tree-based memory network model that is comprised of 3 modules: (1) tree-based patient memory embedding that constructs a personalized and knowledge-enhanced memory network based on  patient's medical histories, (2) trial  embedding that mixes the power of a pre-trained language model with the flexibility of trainable transformer and residual layers to map trial EC sentences into the memory representation space, and (3) beam search based patient memory retrieval that gathers the most relevant medical concepts in the patient memory tree via an efficient top-down exploration guided by the criteria embedding query to offer granular interpretability and also provide targeted input to the final prediction network. 

\subsection*{(1) Dynamic Tree-Based Memory Networks for Personalized Patient Embedding}
To learn a personalized patient embedding, we build a separate memory network for each patient with a distinctive and hierarchical tree structure based on patient's medical history.

\paragraph{Preprocessing} Before accepting any input, we first build a preprocessed mapping such that every medical code in each of the categories of diagnoses, procedures, and medications is mapped to four different hierarchical descriptions and embeddings. For example, the diagnosis code for `Glaucoma secondary to other eye disorders, right eye, mild stage' is mapped to the concept itself and to its parent concepts, including `Glaucoma' and `Diseases of the eye and adnexa.' The embeddings for each description are obtained using a pre-trained ClinicalBERT \cite{ClinicalBERT} model. We run this model from level 1 (describing the broad category of the code) to 4 (representing the specific code itself) descriptions in the Uniform System of Classification (USC) taxonomy of the code and apply MaxPooling on the resulting token embeddings to generate the code embedding.


\paragraph{Building a Personalized Memory Tree} After data preprocessing, we build a memory tree for each patient $P$, whose data are represented as longitudinal EHR information $\mathbf{E} \in \mathbb{R}^{n_v \times 3 \times n_c}$ and its corresponding mask $\mathbf{M}_P \in \mathbb{R}^{n_v \times 3}$, where $\mathbf{E}$ represents a series of patient visits with a possible diagnosis, procedure, and medication code at each visit and $\mathbf{M}_P$ represents whether each modality is present at each visit. For each visit in the EHR, we consider whichever of the diagnosis, procedure, and medication codes that are present according to the mask. For a given code $c$ under consideration, we obtain the four hierarchical descriptions and embeddings from the preprocessed mapping, $[(d_1, \mathbf{e}_1), (d_2, \mathbf{e}_2), (d_3, \mathbf{e}_3), (d_4, \mathbf{e}_4)]$. Looping through each description-embedding pair in order from broad category to granular label, we search for a node in the children of our current base memory tree (the patient's root for the first, level 1 information and the previous level's tree for all subsequent levels) that is keyed on the current description. If such a node is present, we set our new current base memory tree to that node. If it is absent, we add a new tree to our current tree's set of children keyed on the description, and set the new current tree to that. Then, we update the current tree's underlying memory slot $\mathbf{m} \in \mathbb{R}^{n_m}$  using an erase and add scheme according to Eq.~\eqref{eq:update},
\begin{equation}\label{eq:update}
    \mathbf{m}_{new} = \mathbf{m}_{old} * (1 - s \cdot \text{sigmoid}(\mathbf{W}_e\mathbf{e})) + s \cdot \text{tanh}(\mathbf{W}_a\mathbf{e})
\end{equation} 
based on $\mathbf{e}$ for the current embedding and the embeddings of all of the more granular remaining levels, beginning with a scale of $s=1$ and halving $s$ for every additional update to the same cell. We then proceed to the next granularity level. Finally, upon reaching the end of the EHR, we add one final tree node to the children of the patient's root and set its memory cell to the result of a single linear layer used to embed the patient's demographic information $\mathbf{d}$ into the proper memory dimension $\mathbb{R}^{n_m}$. 

\begin{algorithm}[tb]
\caption{Building a Patient's Memory Tree}
\label{alg:MemNet}
\begin{algorithmic}[0]
    \REQUIRE $\mathbf{E}$, $\mathbf{M}_p$, $\mathbf{d}$, preprocessed\_mapping
    \STATE patientTree = Tree( )
    \FOR{$v$ in range(len($\mathbf{E}$))} 
        \FOR{$m$ in $\{diag, proc, med\}$}    
            \IF{$\mathbf{M}_p[v][m] == 1$}
                \STATE code = $\mathbf{E}[v][m]$
                \STATE info = preprocessed\_mapping[code]
                \STATE curNode = patientTree
                \FOR{$l$ in range$(4)$} 
                    \STATE desc, emb = info[$l$]
                    \IF{desc in curNode.children}
                        \STATE curNode = curNode.children[desc]
                    \ELSE
                        \STATE curNode = curNode.addChild(desc)
                    \ENDIF
                    \FOR{$d$ in range$(l, 4)$}    
                        \STATE emb = info[$d$][1]
                        \STATE s = $\frac{1}{2^{l-d}}$
                        \STATE curNode.memory.update(s, emb) // using Eq. \ref{eq:update}
                    \ENDFOR
                \ENDFOR
            \ENDIF
        \ENDFOR
    \ENDFOR
    \STATE patientTree.addChild(``Demographics", demoEmb($\mathbf{d}$))
    \STATE \textbf{return} patientTree
\end{algorithmic}
\end{algorithm}

\begin{figure*}
    \centering
    \includegraphics[scale=0.6]{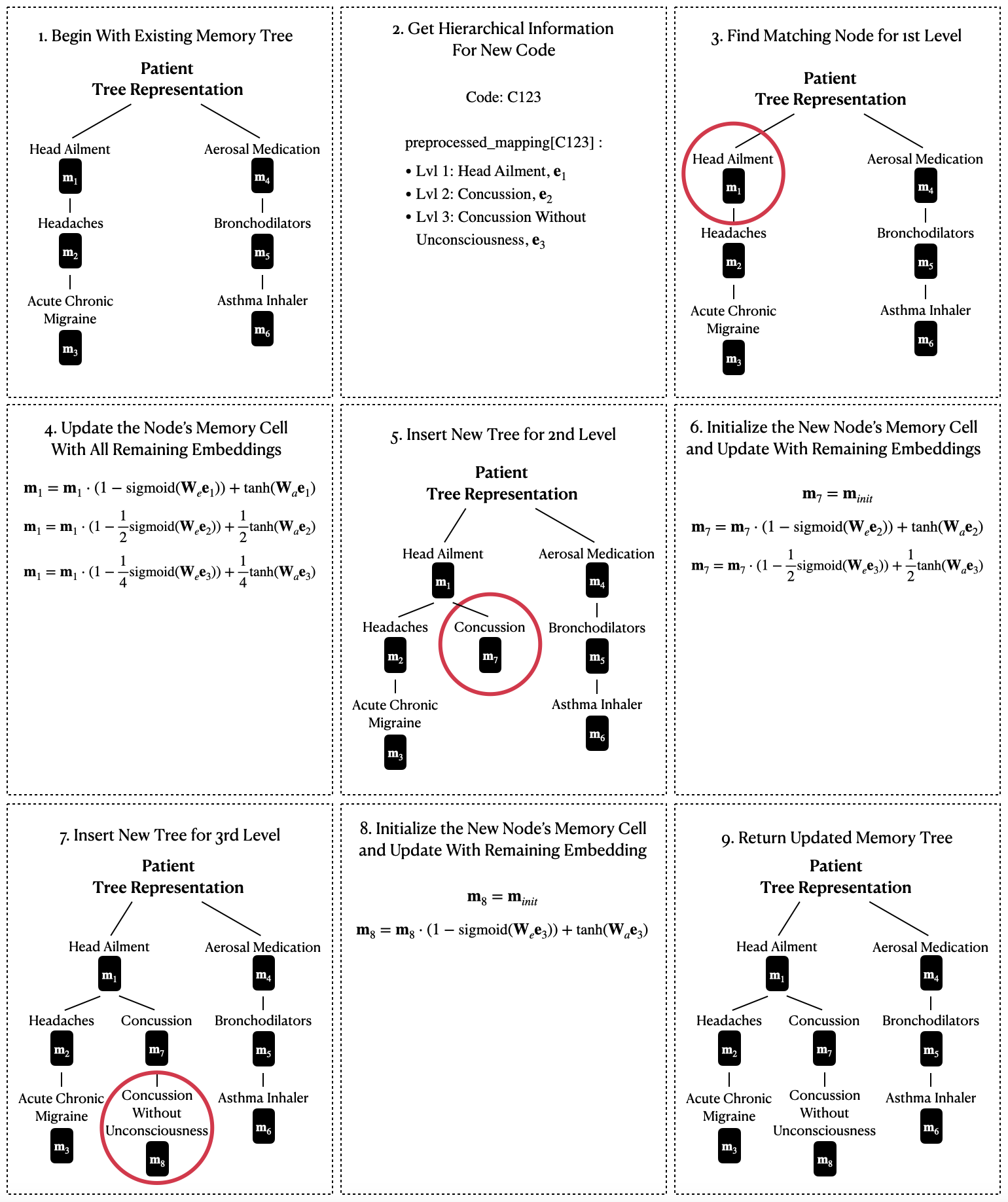}
    \caption{A step-by-step depiction of the insertion of a new code into a patient's memory tree. The same process is then repeated for each code in each visit in the patient's medical record.}
    \label{fig:SampleTree}
\end{figure*}

In this way, we construct a personalized tree-based memory representation for all patients in which the keys to the memory cells have parent-child relationships based on the increasing granularity of the ontology of medical codes from the patient's medical history, and the underlying cells themselves are a function of the embedding of the cell's key, the embeddings of all its children's keys, and the temporal nature of the sequence of visits. This tree is personalized and flexible such that only those relatively few medical codes in a patient's history appear and are used to represent them, maintaining the granular precision and interpretability of individual codes without producing an overly large and unwieldy structure containing each of the tens of thousands of possible codes. We summarize the construction procedure in Algorithm \ref{alg:MemNet} and provide a step-by-step depiction of the insertion of a new code into a patient's memory tree in Figure \ref{fig:SampleTree}.

\paragraph{Efficiency} While this patient representation approach is more granular and personalized, it is also more efficient. In fact, the lack of a need for static memory slots or embedding mappings for each code individually means that the number of parameters of \method's patient representation module is less than half of that of leading baselines. Specifically, our implementation of \method's memory tree mechanism has just 199,299 parameters as compared to 497,280 in \cite{COMPOSE} and 571,904 in \cite{DeepEnroll}, respectively.

\subsection*{(2) Trial Criteria Embedding}
Given a criteria sentence $\mathbf{S} \in \mathbb{R}^{n_s \times n_w}$ represented as a sequence of one-hot vectors for each token and its mask $\mathbf{m}_s$, we embed it into a vector format that we can use to query our memory trees. We first feed the sequence through the same pre-trained ClinicalBERT \cite{ClinicalBERT} model as in the previous module to produce meaningful, enriched embeddings for each token, drawing on pre-training from a much larger dataset than we may have access to. We then use that sequence of embeddings, $\mathbf{S}_e^{(1)}$ as the input to the trainable part of our network. We feed it through two additional Transformer encoder layers (from the original transformer architecture proposed in \cite{Attention}) that further tune the embeddings of each token for use in our specific problem setting and produce $\mathbf{S}_e^{(2)}$ and $\mathbf{S}_e^{(3)}$ respectively. We then perform MaxPooling over the final sequence of embeddings to compress our representation to a single embedding, $\mathbf{s}_e^{(1)}$, regardless of the sentence's length. Finally, we pass this single embedding through two residual blocks to produce the module's output, $\mathbf{q} \in \mathbb{R}^{n_m}$, a vector of the same memory dimension from the previous module. 
In this way, we can embed a sentence into $\mathbb{R}^{n_m}$ as our query in the next module.

\subsection*{(3) Attentional Beam Search Based Retrieval}
The third module of \method uses the criteria embedding $\mathbf{q}$ as a query to retrieve patient memories. Specifically, for any given memory cell associated with a node in our memory tree, we first calculate its attention score via Eq.~\eqref{eq:attention},
\begin{equation}\label{eq:attention}
    a_i = \mathbf{q}^T\mathbf{m}_i
\end{equation}

To perform the beam search, we keep a running list of our current ``beam" in which we store triplets of memory cells, attention scores, and booleans stating whether the cell has already been expanded. We start with a list containing just the root of the memory tree, and generate a new list of all  children of the unexpanded cells in our beam (setting each of their booleans to true as we do). For every cell in this new list, we calculate their attention scores, and add triplets of cell, attention score, and false to our beam. We sort the beam by attention score and keep the top $b$ triplets, proceeding to the next iteration. Once finishing the loop, we take a softmax over the final $b$ attention scores and return a weighted sum of the memory cells as the response $\mathbf{r} \in \mathbb{R}^{n_m}$ to the query. 

This beam search algorithm allows us to parse through a patient's variable-length and possibly extensive medical history in a focused and efficient manner, proceeding down the promising paths in the tree (making use of the fact that each cell is a function not only of its own medical meaning but also of those of its descendants) while ignoring irrelevant information. 
We summarize the beam search in Algorithm \ref{alg:BeamQuery}.

\begin{algorithm}
\caption{Attentional Beam Search of the Memory Tree}
\label{alg:BeamQuery}
\begin{algorithmic}[0]
    \REQUIRE PatientTree, $\mathbf{q}$, $b$
    \STATE beam = [(PatientTree, $-\infty$, False)]
    \STATE // beam is a list of triplets t where t[0] is a tree node, t[1] is the node's attention, and t[2] says if the node is expanded
    \WHILE{remainsUnexpanded(beam)}
        \STATE childBeam = [ ]
        \FOR{t in beam}
            \IF{t[2] == False}
                \STATE childBeam.extend(t[0].children)
                \STATE t[2] = True
            \ENDIF
        \ENDFOR
        \FOR{c in childBeam}
            \STATE att = $\mathbf{q}^T \cdot$ c.memory
            \STATE beam.append((c, att, False))
        \ENDFOR
        \STATE beam = beam.sort()[0$\,:\,$$b$]
    \ENDWHILE
    \STATE attentions = softmax([t[1] for t in beam])
    \STATE memories = [t[0].memory for t in beam]
    \STATE \textbf{return} sum(attentions $*$ memories)
\end{algorithmic}
\end{algorithm}

\subsection*{(4) Output and Explanation Tree}
Our architecture then concludes with a final network that maps the concatenation of the criteria embedding query and the memory network's response to the final probabilities via a series of fully connected layers as in Eq.~\eqref{eq:prob},
\begin{align}\label{eq:prob}
\begin{split}
    \mathbf{h}_1 &= \text{concat}(\mathbf{q}, \mathbf{r}) \\
    \mathbf{h}_2 &= \text{ReLU}(\mathbf{h}_1\mathbf{W}_1 + \mathbf{b}_1) \\
    \mathbf{h}_3 &= \text{ReLU}(\mathbf{h}_2\mathbf{W}_2 + \mathbf{b}_2) \\
    \hat{\mathbf{y}} &= \text{softmax}(\mathbf{h}_3\mathbf{W}_3 + \mathbf{b}_3)
\end{split}
\end{align}
where $\mathbf{W}_1 \in \mathbb{R}^{256 \times 128}$, $\mathbf{b}_1 \in \mathbb{R}^{128}$, $\mathbf{W}_2 \in \mathbb{R}^{128 \times 32}$, $\mathbf{b}_2 \in \mathbb{R}^{32}$, $\mathbf{W}_3 \in \mathbb{R}^{128 \times 3}$, and $\mathbf{b}_3 \in \mathbb{R}^{3}$. $\hat{\mathbf{y}} \in \mathbb{R}^3$ is the model's final output and can be used as its prediction.

The model is then trained using the  loss function introduced by \cite{COMPOSE}, which is the sum of a typical cross-entropy loss component for the core classification problem with an additional inclusion/exclusion loss component $\mathcal{L}_D$  designed to align the retrieved patient memory embedding with inclusion criteria embeddings while maximizing the distance between those same patient representations and exclusion criteria embeddings. The loss function is given by Eq.~\eqref{eq:loss},
\begin{equation}\label{eq:loss}
    \mathcal{L}_D = \left\{
        \begin{array}{ll}
            1 - d(\mathbf{r},\mathbf{q}) & C \text{ is IC}  \\
            max(0, d(\mathbf{r},\mathbf{q}) - \alpha) & C \text{ is EC}
        \end{array}
    \right.
\end{equation}
where IC stands for inclusion criterion, EC stands for exclusion criterion, $d$ calculates the distance between two embeddings, and $\alpha$ is a hyperparameter which sets a minimum learned distance between two embeddings. We then backpropgate the sum of this and the cross-entropy loss component through our architecture to train our model.

\paragraph{Explanation Tree} In addition to matching results, \method also outputs an interpretable tree structure to explain and verify its predictions. The memory tree built to represent the patient and the attention paid to different individual medical codes during the beam query combine to form a decision tree. Not only can the most relevant medical codes and descriptions be provided in a brief list for immediate verification, a manual review of the tree can follow the flow of attention down the tree to map the logic underlying the decision making process from broad medical categories to specific diagnoses, procedures, or medications that satisfy or fail to meet the given criteria.

\section{Experiment}

\begin{table*}
\centering
\caption{Test Set Matching Results}
\begin{subtable}{0.62\linewidth}
    \centering
    \caption{Criteria Level Results}
    \begin{tabular}{c|cccc} \toprule
        & Accuracy & F1 Score \\ \midrule
        DeepEnroll    & $0.9040 \pm 0.0009$ & $0.9169 \pm 0.0008$ \\
        COMPOSE    & $0.9529 \pm 0.0006$ & $0.9589 \pm 0.0005$  \\
        \method     & $\mathbf{0.9561 \pm 0.0006}$ & $\mathbf{0.9620 \pm 0.0005}$  \\\bottomrule
    \end{tabular}
    \label{tab:CriteriaLevel}
\end{subtable}%
\begin{subtable}{0.38\linewidth}
    \centering
    \caption{Trial Level Results}
    \begin{tabular}{c|c} \toprule
        & Accuracy \\ \midrule
        Criteria2Query    & $0.614^*$ \\
        DeepEnroll    & $0.766 \pm 0.001$ \\
        COMPOSE    & $0.847 \pm 0.001$ \\
        \method     & $\mathbf{0.849 \pm 0.001}$ \\\bottomrule
    \end{tabular}
    \label{tab:TrialLevel}
\end{subtable}%
\label{tab:ClassificationResults}
\end{table*}

\subsection{Experimental Setup}
We provide more details regarding datasets, model hyperparameters, model training, and evaluation techniques in the appendix. Below are some key information.

\paragraph{Datasets}  (1) \textbf{Clinical trial data}: We utilize 590 randomly selected clinical trials from the publicly available ClinicalTrials.gov database and extract the inclusion-exclusion criteria from each for a total of 12,445 criteria sentences. (2) \textbf{Patient data}: We extract EHR from a large health data company's proprietary claims data for 83,371 patients from 2002 to 2018, each of whom is a match to at least one of the previously selected trials. Those EHRs are encoded into the sequential format which is passed into the model, consisting of a series of visits which each may have a diagnosis, procedure, and medication code and whose codes belong to the Uniform System of Classification system (USC) with four ontological levels. 

We label each criterion and its corresponding patient(s) as either ``match" or ``mismatch" depending on whether the criteria was for inclusion to or exclusion from the trial, and we randomly sample one inclusion and exclusion criteria from another trial to assign the third label of ``unknown." This process yields  397,321 labeled pairs in the final dataset.

\paragraph{Baselines}
We consider the following baseline models.
\begin{enumerate}
    \item{\textbf{Criteria2Query}~\cite{Criteria2Query} parses criteria statements into SQL queries that can be run against structured patient data translate them to a set of structured attributes. And then use
these attributes to identify patient cohorts.}
    \item{\textbf{DeepEnroll}~ \cite{DeepEnroll} utilizes a pair of embedding networks to embed both the patient and criteria statement into a shared representation space and then uses an alignment matrix to predict the matching results.}
    \item{\textbf{COMPOSE}~\cite{COMPOSE} utilizes the hierarchical information that accompanies medical codes to build a memory network patient representation. The criteria statement is then embedded and used as a query to that memory network whose response is concatenated with the embedding and fed into the final response network for matching.\\}
\end{enumerate}

\paragraph{Evaluation Strategy and Metrics} We use both F1 score and Accuracy to measure the criteria level performance, and use Accuracy to measure trial level performance.
We fix a test set of $30\%$ patients, and divide the rest of the dataset into  training and validation with a proportion of $90\%:10\%$. We fix the best model on the validation set and report the performance on the test set. We bootstrap the test set 1,000 times and report both mean and 95\% confidence interval for all results.

\begin{table*}
\centering
    \caption{Trial Level Matching Accuracy by Trial Type}
    \begin{tabular}{c|ccc|ccc} \toprule
        & \multicolumn{3}{c|}{Trial Disease Type} & \multicolumn{3}{c}{Trial Phase} \\
        & Diabetes   & Parkinson's   & Rare   & Phase 1   & Phase 2   & Phase 3 \\ \midrule
        Criteria2Query     & - & - & -   & $0.302^*$ & $0.643^*$ & $0.587^*$ \\
        DeepEnroll     & $0.380 \pm 0.001$ & $0.463 \pm 0.001$ & $0.865 \pm 0.001$   & $0.373 \pm 0.001$ & $0.835 \pm 0.001$ & $0.720 \pm 0.001$ \\
        COMPOSE    & $0.619 \pm 0.001$ & $0.551 \pm 0.002$ & $0.941 \pm 0.001$   & $0.397 \pm 0.001$ & $\mathbf{0.920 \pm 0.001}$ & $0.828 \pm 0.001$ \\
        \method    & $\mathbf{0.631 \pm 0.001}$  & $\mathbf{0.632 \pm 0.001}$  & $\mathbf{0.970 \pm 0.001}$   & $\mathbf{0.402 \pm 0.002}$ & $0.916 \pm 0.001$ & $\mathbf{0.837 \pm 0.001}$\\\bottomrule
    \end{tabular}
    \label{tab:TrialType}
\end{table*}

\subsection{Results} 

We design experiments to answer the following questions.
\begin{enumerate}
    \item Does personalized patient data embedding of \method improve the performance of patient-trial matching?
    \item Does \method provide interpretable results?
\end{enumerate}

\subsection*{Q1. Performance Analysis for Patient Trial Matching}

\begin{figure*}
\centering
\includegraphics[scale=0.42]{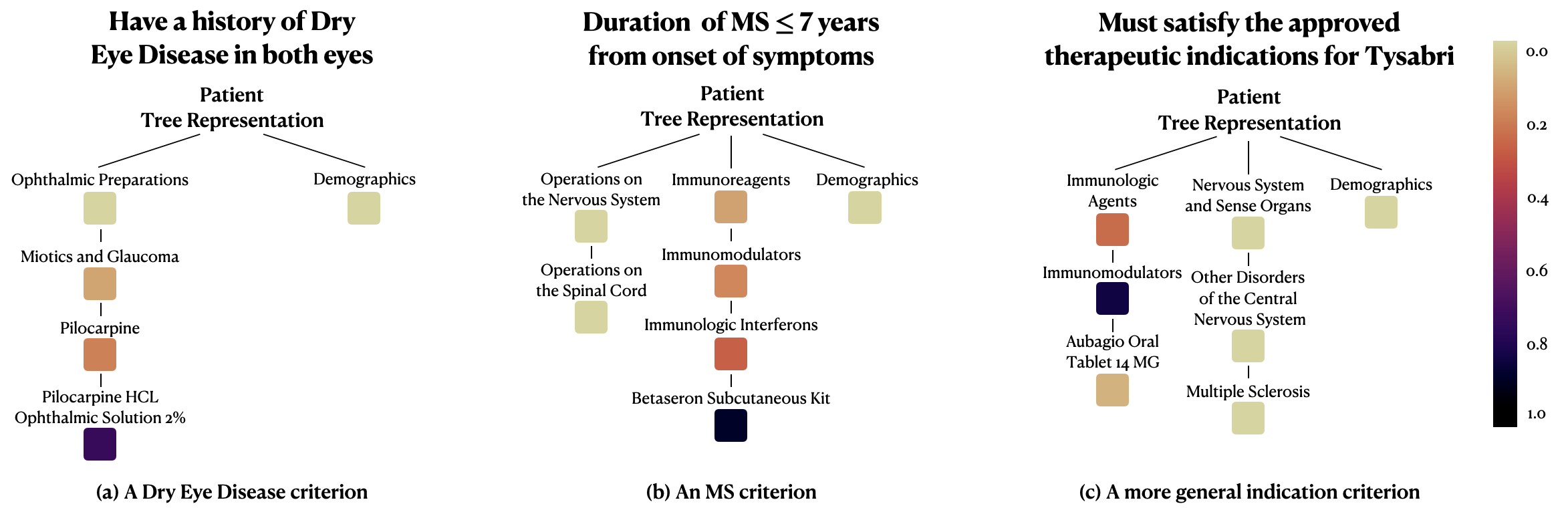}
\caption{Three case studies visualizing the attention to different cells in a patient's memory tree in response to an eligibility criterion query to evaluate its underlying reasoning and interpretability. The numbers are the attention scores. }
\label{fig:CaseStudies}
\end{figure*}

\begin{figure}
\centering
\includegraphics[scale=0.36]{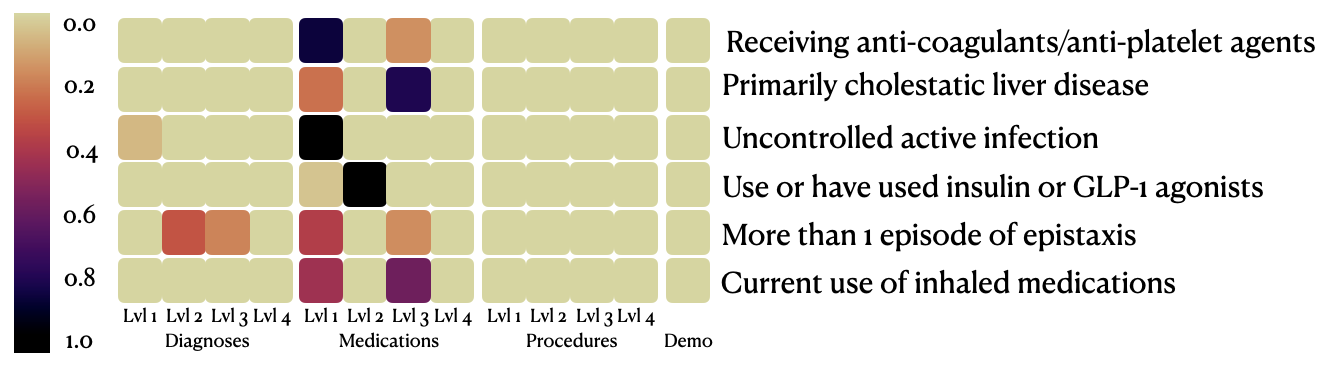}
\caption{Case studies visualizing the attention to different memory cells for COMPOSE in response to six different criteria to demonstrate its inconsistent and insufficient interpretability.}
\label{fig:COMPOSE_Interpretability}
\end{figure}

In Table \ref{tab:ClassificationResults}, we show the model performance for both criteria level and trial level matching. For criteria level results in Table \ref{tab:CriteriaLevel}, \method outperforms all baselines including the state-of-the-art COMPOSE model. It achieves 96.20\% F1 Score and 95.61\% accuracy, showing that \method can accurately match patients with individual criteria statement.

For trial level comparison, we want to see if \method can correctly predict all criteria statement for a given trial. Specifically, for all patients in our test set, for each trial that they were enrolled in, we aggregate all of the eligibility criteria statements for that trial and count the model as correctly matching the patient to the trial if it labels all inclusion criteria statements as a ``match" and all exclusion criteria statements as a ``mismatch." We report the accuracy of such evaluations in Table \ref{tab:TrialLevel}. Here we again see that \method is able to outperform all baselines, correctly enrolling 84.9\% of patients to their respective trials.

\paragraph{Trial Level Comparison by Disease and Phase}
Furthermore, we would like to explore the model performance for trials of different diseases and phases as the performance could vary significantly.
We list some results in Table \ref{tab:TrialType}. Here we see \method significantly outperforms all baselines within each of the disease categories of Diabetes, Parkinson's Disease, and Rare Disease trials as well as phase 1 and phase 3 trials, closely approximating COMPOSE's performance in phase 2 trials. 

For all models, the performance matching patients with diabetes and Parkinson's Disease (PD) trials are lower than rare diseases. This is due to patients with diabetes and PD often having complex conditions with heterogeneous manifestation. So, the criteria for these diseases often use more general descriptions that are more difficult to match. Alternately, rare diseases typically have more distinct conditions and 
thus rely on more homogeneous and distinct lower level concepts.

We see a similar divide in overall performance at the phase level where the performance on Phase 1 trials is much lower for each compared model. This is due to the fact that those trials test the safety of new drugs and so typically enroll a limited number of patients who have exhausted other treatment options. As such, they make up a much smaller fraction of our training data and it therefore makes sense that the models perform worse on those types of data.

\subsection*{Q2. Analysis of Interpretability}
In addition to strong prediction power, interpretability is also essential to adopt the model results in real-world settings. Therefore, in this section we attempt to evaluate the interpretability of \method against the current state-of-the-art COMPOSE model. 

To evaluate, we look at three case studies for \method and examine the underlying tree-based memory structure and the attention each code was given for the corresponding eligibility statement to verify that this process and its interpretability are working as intended. We also contrast our tree-structure model explanation with the attention heatmap generated by the COMPOSE~\cite{COMPOSE} model. We do not consider DeepEnroll~\cite{DeepEnroll} as it does not provide any explanation mechanism.

In the first case study, shown in Figure \ref{fig:CaseStudies} (a), our model correctly attends to a specific eye drop medication in response to a Dry Eye Disease criterion statement. In the second case study in Figure \ref{fig:CaseStudies} (b), our model similarly attends to a specific Multiple Sclerosis medication for an MS criterion. Finally, for the third case study in Figure \ref{fig:CaseStudies} (c), our model attends to the relevant higher level medical concepts in response to a more general therapeutic indication criterion. 


As a comparison, we show the attention output of COMPOSE in Figure \ref{fig:COMPOSE_Interpretability}. There we see that while COMPOSE is typically attending to the correct medical categories such as with medications for the inhaled medications criterion and diagnoses for the epistaxis criterion, those broad categories of attention do not provide true interpretability and can not be easily verified in a real-world setting.

To summarize, \method shows the ability to provide correct and easily verifiable interpretability in response to a diverse set of eligibility criteria. This granular level of interpretability makes \method fit for real world clinical trial recruitment settings.


\section{Conclusion}
We propose a new personalized dynamic tree-based memory network model, named \method, for interpretable patient trial matching. Our method uses a dynamic tree-based memory network that takes advantage of the hierarchical ontology of medical codes and is fully personalized for each individual patient. It also uses an attentional beam search based patient memory retrieval to gather the most relevant medical concepts in the patient memory tree for accurate and interpretable patient-trial matching.
We demonstrate the strong matching accuracy and interpretability of \method through performance on a real-world test set and a series of case studies.


\bibliographystyle{ACM-Reference-Format}
\bibliography{refs}

\appendix

\section{Dataset Availability and Limitations}
\label{sec:Dataset}
Here we briefly discuss limitations to our experiment in terms of the dataset. While we explored additional options, the dataset we used was the only available real-world dataset that we were able to identify, and we chose not to utilize low-quality synthetic datasets. This real-world dataset is furthermore proprietary to a clinical trial operations company and so unable to be shared. However, we do share limited data samples to demonstrate its precise data formatting, all of our preprocessing scripts, and the publicly available trial half of the dataset. 

Additionally, despite making large advancements in terms of interpretability while simultaneously achieving state of the art effectiveness, our \method model (as well as the baselines it is compared against) is limited by the contents of this dataset as well. For example, many eligibility criteria that \method fails to accurately classify require either demographic information that is not present in our dataset (such as pregnancy information) or exact rather than relative temporal information that isn't represented either. So, to further improve performance, our dataset itself would need to be improved. 

However, we feel that these limitations only further demonstrate the novelty of the task and the need for this line of research given the importance of improving clinical trial enrollment. We believe that this paper is necessary to spur additional research to hopefully inspire the creation of other benchmark datasets and methods in the future.

\section{Model Training, Validation, and Code} \label{sec:ExperimentalDetails}
Next, we discuss specific details regarding our model and experimental setup, beginning with model and training hyperparameters. We use a 0.7-0.3 training-test split with an additional 0.9-0.1 training-validation split during training. We use the Adam optimizer with learning rate 0.0001 (which was arrived upon via experimentation of interpolated values from 0.01 to 0.00001). We begin with a beam size of 25 in order to allow the model to learn before linearly decreasing to 4 by the end of training. We use a batch size of 384 (as it is the largest we can fit into memory) and train for 20 epochs, which takes roughly a week. 

While training each of our compared models takes too long to reasonably bootstrap the models themselves, we generate 95\% confidence intervals for each result through bootstrapping the test sets 1000 times. Finally, we implement all of the models and train using the PyTorch framework \cite{PyTorch} via a single NVIDIA GeForce RTX 3090 GPU. We provide all of our source code at \url{https://anonymous.4open.science/r/TREEMENT/}.

\begin{table*}[]
    \centering
    \caption{Specific disease names that make up each disease category}
    \begin{tabular}{c|p{10.5cm}} \toprule
        Category & \multicolumn{1}{c}{Diseases} \\ \midrule
        Diabetes & Diabetes Mellitus, Type 1; Diabetic Retinopathy; Diabetes; Diabetes Mellitus; Diabetes Mellitus, Type 2; Type 1 Diabetes Mellitus; Painful Diabetic Neuropathy; Type 2 Diabetes; Type 2 Diabetes Mellitus; Diabetic Peripheral Neuropathy; Diabetic Nephropathy; Diabetic Macular Edema\\ \midrule
        Parkinson's Disease & Advanced Parkinson's Disease; Early Stage Parkinson Disease; Early Stage Parkinson's Disease; Parkinson Disease; Parkinson's Disease; Advanced Stage Parkinson's Disease; Parkinson's Disease Psychosis; Idiopathic Parkinson's Disease; Advanced Idiopathic Parkinson's Disease\\ \midrule
        Rare Diseases & Friedreich Ataxia; Advanced Gastric Cancer; IgA Nephropathy; Non-Squamous Non-Small Cell Lung Cancer; Hereditary Angioedema (HAE); ALS Caused by Superoxide Dismutase 1 (SOD1) Mutation; Hemophilia A With Inhibitors; Neuromyelitis Optica; Hereditary Thrombotic Thrombocytopenic Purpura (TTP); Huntington's Disease; Progressive Multifocal Leukoencephalopathy; Fabry Disease; Hilar Cholangiocarcinoma; Immune Reconstitution Inflammatory Syndrome; Spinal Muscular Atrophy; Aggressive Systemic Mastocytosis; Anxiety Neuroses; Dravet Syndrome; Refractory Generalized Myasthenia Gravis; Leber's Hereditary Optic Neuropathy (LHON); Fabry Disease; Lennox-Gastaut Syndrome; Immune Reconstitution Inflammatory Syndrome\\ \bottomrule
    \end{tabular}
    \label{tab:Diseases}
\end{table*}

\section{Trial Type Information}
We provide specific details on our trial type splits. The split by trial phase is straightforward. In our test set, we have 1,012 Phase 1 trial-patient pairs, 9,945 Phase 2 trial-patient pairs, and 11,311 Phase 3 trial-patient pairs. The split by trial disease is performed by checking whether the listed disease for each trial in the test set falls into one of our three disease categories based on a doctor-curated selection of diseases for each category. The specific diseases included in each category can be found in Table \ref{tab:Diseases}. In our test set, we have 785 Diabetes trial-patient pairs, 819 Parkinson's Disease trial-patient pairs, and 3,177 Rare Disease trial-patient pairs.

\end{document}